\documentclass{article}

\usepackage{arxiv}

\usepackage[utf8]{inputenc} 
\usepackage[T1]{fontenc}    
\usepackage{hyperref}       
\usepackage{url}            
\usepackage{booktabs}       
\usepackage{amsfonts}       
\usepackage{nicefrac}       
\usepackage{microtype}      
\usepackage{graphicx}
\usepackage{natbib}
\usepackage{doi}

\usepackage{amssymb}
\usepackage{latexsym}

\usepackage{amsmath}
\usepackage{wasysym}
\usepackage{booktabs}

\usepackage{graphicx}
\usepackage{array}
\usepackage{mathtools}
\usepackage{multirow}

\usepackage{caption}
\usepackage{subcaption}

\usepackage{authblk}

\title{Automatic registration with continuous pose updates for marker-less surgical navigation in spine surgery}

\author[1,2]{Florentin Liebmann}
\author[1,2]{Marco von Atzigen}
\author[3]{Dominik Stütz}
\author[4]{Julian Wolf}
\author[5]{Lukas Zingg}
\author[5]{Daniel Suter}
\author[1]{Laura Leoty}
\author[1]{Hooman Esfandiari}
\author[2]{Jess G. Snedeker}
\author[3,6]{Martin R. Oswald}
\author[3,7]{Marc Pollefeys}
\author[5]{Mazda Farshad}
\author[1]{Philipp Fürnstahl}

\affil[1]{Research in Orthopedic Computer Science, Balgrist University Hospital, University of Zurich, Zurich, Switzerland}
\affil[2]{Laboratory for Orthopaedic Biomechanics, ETH Zurich, Zurich, Switzerland}
\affil[3]{Computer Vision and Geometry Group, ETH Zurich, Zurich, Switzerland}
\affil[4]{Product Development Group, ETH Zurich, Zurich, Switzerland}
\affil[5]{Department of Orthopedics, Balgrist University Hospital, University of Zurich, Zurich, Switzerland}
\affil[6]{Computer Vision Lab, University of Amsterdam, Amsterdam, Netherlands}
\affil[7]{Microsoft Mixed Reality and AI Zurich Lab, Zurich, Switzerland}

\renewcommand{\shorttitle}{Automatic registration for marker-less surgical navigation in spine surgery}

\begin{document}

\maketitle

\begin{abstract}
Established surgical navigation systems for pedicle screw placement have been proven to be accurate, but still reveal limitations in registration or surgical guidance. Registration of preoperative data to the intraoperative anatomy remains a time-consuming, error-prone task that includes exposure to harmful radiation. Surgical guidance through conventional displays has well-known drawbacks, as information cannot be presented in-situ and from the surgeon's perspective. Consequently, radiation-free and more automatic registration methods with subsequent surgeon-centric navigation feedback are desirable. In this work, we present an approach that automatically solves the registration problem for lumbar spinal fusion surgery in a radiation-free manner. A deep neural network was trained to segment the lumbar spine and simultaneously predict its orientation, yielding an initial pose for preoperative models, which then is refined for each vertebra individually and updated in real-time with GPU acceleration while handling surgeon occlusions. An intuitive surgical guidance is provided thanks to the integration into an augmented reality based navigation system. The registration method was verified on a public dataset with a mean of 96\% successful registrations, a target registration error of 2.73 mm, a screw trajectory error of 1.79° and a screw entry point error of 2.43 mm. Additionally, the whole pipeline was validated in an ex-vivo surgery, yielding a 100\% screw accuracy and a registration accuracy of 1.20 mm. Our results meet clinical demands and emphasize the potential of RGB-D data for fully automatic registration approaches in combination with augmented reality guidance.
\end{abstract}


\section{Introduction}
Complex orthopedic procedures, such as pedicle screw placement, can benefit from computer assistance in regards to safety and accuracy (\cite{gelalis2012accuracy,perdomo2019accuracy}). Nevertheless, computer-assisted orthopedic surgery (CAOS) only accounts for an estimated 5\% of all orthopedic surgeries performed in North America, Europe and Asia (\cite{joskowicz2016computer}). Many state-of-the-art navigation systems for pedicle screw placement comprise three main components: planning, registration and navigation. The latter two strongly contribute to the low clinical adoption (\cite{hartl2013worldwide,nadeau2015qualitative,joskowicz2016computer}). Existing registration approaches tend to be time-consuming, cumbersome and often involve radiation. This hinders real-time application for surgical navigation, which itself suffers from limitations caused by conventional visualization techniques.

State-of-the-art CAOS systems commonly require intraoperative imaging or manual anatomy digitization by the surgeon for registration (\cite{markelj2012review}). Both are time-consuming processes. In addition, except for ultrasound, intraoperative imaging, e.g. fluoroscopy or cone-beam CT, always comes along with bulky equipment and radiation exposure for patient as well as OR personnel. Despite a multitude of techniques for 2D/3D registration (\cite{sundar2006novel,esfandiari2019comparative,miao2016cnn}), which reduces radiation, registration failures are an existing problem even when reference markers are used (\cite{zhang2019risk}).

Besides the registration issue, most state-of-the-art navigation systems provide visualizations on 2D monitors in the OR periphery, which can cause attention shift and may increase the cognitive load for the surgeons, e.g. hand-eye coordination (\cite{brendle2020can,Qian2017,leger2017quantifying}).Given the recent advent of augmented reality (AR) solutions and their potential in the realm of medicine, their application in intraoperative settings to provide surgical guidance should be considered in hopes of alleviating the aforementioned limitations (\cite{eckert2019augmented,birlo2022utility}). The use of AR for spine surgery, and pedicle screw placement in particular, has been investigated thoroughly in the past few years, showing the benefits that the technology could bring to the field (\cite{Ma2017,elmi2019pedicle,Gibby2019,liebmann2019pedicle,molina2019augmented,farshad2021operator,farshad2021first,liu2021clinical,uddin2021augmented,von2022marker}). While there are mitigation strategies tailored to the visualization challenges of the exsiting CAS solutions (e.g. \cite{wolf2023different}), the question about an ideal registration remains.

As a potential remedy for the aforementioned registration difficulties, one approach towards a more automatic and radiation-free registration is the intraoperative 3D reconstruction of the target anatomy using depth sensing hardware and the associated computer vision software. \cite{ji2015patient} used two co-calibrated RGB cameras mounted to a surgical microscope for registration of preoperative spine CT images in a clinical trial. In their work, the authors pursued a semi-automatic segmentation approach to localize the anatomy of interest in the 2D RGB images based on manual surgeon annotations and region growing followed by a 3D reconstruction module. They reached a registration accuracy of 1.43 mm, but the manual annotations make real-time use in a surgical setting cumbersome. The path of stereo feature matching in open spine surgery was further demonstrated to be promising in recent work by \cite{manni2020towards}, who achieved a better than 0.5 mm 3D triangulation error on grayscale images as evaluated on data from 23 patients. Besides methods that have been reported and evaluated in academic settings, a commercial navigation system for surface-reconstructing, radiation-free spine surgery navigation is available that operates based on a structured light sensor integrated into OR lamps (7D Surgical Inc., Toronto, ON, Canada, \cite{faraji2020machine}. The authors have reported the registration time to be less than 20 s, and re-registration (in case of perceived registration inaccuracies) to be even faster. However, the lamp-integrated hardware is not versatile. Furthermore, registration starts with manual point sampling and has to be performed for each vertebral level individually. Motion detection and compensation relies on markers clamped to the anatomy, which is the standard approach. While such techniques can diminish some of the concerns related to radiation exposure, time and cost of a common registration pipeline, they still require a residual registration process between the reconstructed anatomy and the patient, making them susceptible to issues such as sub-optimal manual input, poor initialization and small capture range. More recent algorithms have looked into achieving a higher level of registration autonomy through the utilization of artificial intelligence (AI) concepts. In \cite{felix2021towards}, an RGB-D sensor was used for automatic registration of preoperative femural and tibial 3D models to intraoperative cadaveric anatomy. The RGB images were segmented with a neural network allowing for the corresponding 3D segmentation of the reconstructed anatomy. The reconstructed 3D models were then automatically registered to the anatomy using a RANSAC-based method. Through the analysis of the results, the authors have reported that a considerable part of the error was attributed to the infrared-based RGB-D sensor, an observation that was affirmed in other studies (e.g., \cite{gu2021feasibility}). \cite{hu2022automatic} investigated a femur registration and tracking approach using point cloud data that finds a global alignment between a peroperative 3D reference model and intraoperative depth camera data based on RANSAC, followed by an iterative closest point (ICP) refinement. A PointNet-based network was proposed in this study to restore the surface of the unmodified bone captured by the depth camera before using it for registration, coping intraoperative bone surface modification. The employment of the network reduced the registration RMSE from 2.40 mm to 2.07 mm, but the improvement did not reflect significantly on the pose error when compared to a ground truth tracking. An early prototype of a complete AR-based navigation approach using an optical see-through head-mounted display (HMD) for total shoulder arthroplasty evaluated on synthetic bone models was presented in \cite{gu2021calibration}. The 3D model counterpart of the synthetic bone model is first aligned manually as a movable AR rendering, followed by an ICP refinement. Thereby, the intraoperative data was a point cloud, originating from a co-calibrated external RGB-D sensor. The point cloud was computed from a disparity map which was generated by a tansformer-based network with the two RGB images of the depth sensor as input. After registration, an additional fiducial marker clamped to the anatomy is responsible for motion compensation. An average pin placement accuracy of 4.66° and 3.8 mm was achieved. These recent research contributions show that AI-based algorithms using surface data, potentially in combination with RGB, could advance registration approaches in CAOS. What stands out is that there remains a dependency on a coarse initial alignment as well as reference markers for motion detection and/or compensation.

The goal of this study was to tackle the aforementioned drawbacks of the current navigation systems by developing an efficient, radiation-free and real-time approach for automatic registration of intraoperative RGB-D data to the underlying patient with the potential downstream goal of providing a more accurate and faster pedicle screw placement alternative under AR guidance. Our method comprises of a registration module for automatic piecewise registration of preoperative lumbar spine 3D models to intraoperative RGB-D data with pose updates during surgical interaction as well as a navigation module for AR-guided pedicle screw placement. The registration module was developed and evaluated on the public SpineDepth dataset (\cite{liebmann2021spinedepth}) of pose-annotated cadaveric surgery RGB-D recordings. Data collected from simulated pedicle screw placement interventions was used to evaluate the registration success. Finally, the full pipeline (registration + navigation) was validated in an \textit{ex-vivo} setup, where a surgeon placed ten pedicle screws in a cadaveric lumbar spine under AR guidance.

\section{Material and methods}
\label{reg-sec:methods}
\begin{figure}[htb!]
\centering
\includegraphics[width=\textwidth]{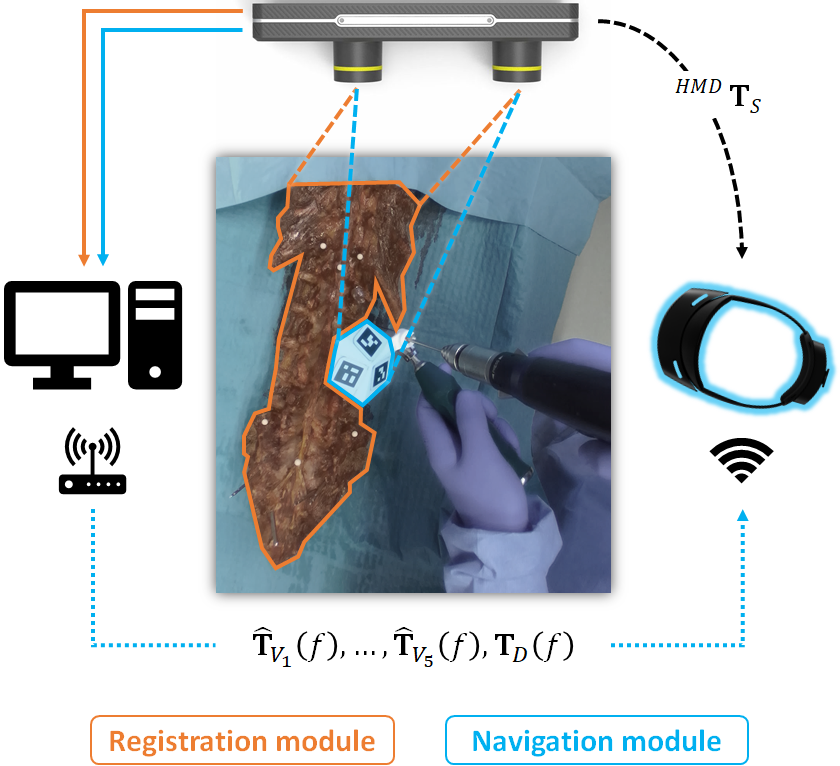}
\caption{Setup overview. Solid lines denote a wire, dotted lines denote a wireless connection and dashed lines denote a transformation. The RGB-D sensor observes the surgical site from the top and serves as input for the registration and navigation module. The former outputs five poses, one for each vertebra $V_{i}$ and frame $f$: $\hat{\textbf{T}}_{V_{1}}(f),\dots,\hat{\textbf{T}}_{V_{5}}(f)$. The latter tracks the surgical drill sleeve and outputs its pose $\textbf{T}_{D}$ for each frame: $\textbf{T}_{D}(f)$. All poses are streamed to the HMD wirelessly. The second part of the navigation module comprises AR guidance for pedicle screw placement, based on the received poses. ${}^{HMD}\textbf{T}_{S}$ denotes the transformation between the RGB-D sensor's coordinate frame and the one of the HMD. It is found using standard chessboard detection after the RGB-D sensor has been positioned.}
\label{reg-fig:setup}
\end{figure}

The proposed hardware setup consists of the following components: an RGB-D sensor, a GPU-enabled workstation and a HMD. In our case, a ZED Mini (Stereolabs Inc., San Francisco, CA, USA), a HP Z2 (HP Inc., Palo Alto, CA, USA) with an Nvidia GeForce RTX 2080 SUPER (Nvidia Corporation, Santa Clara, CA, USA) and a Microsoft HoloLens 2 (Microsoft Corporation, Redmond, WA, USA) were used (Fig. \ref{reg-fig:setup}).

The main contribution of this work lies in the registration and navigation method. It can be subdivided into two modules: the registration module and the navigation module (Fig. \ref{reg-fig:setup}). The RGB-D sensor observes the surgical site from the top and serves as input for both modules. The registration module (Section \ref{reg-sec:registration}) is responsible for the automatic segmentation and registration of lumbar vertebrae L1--L5 and outputs five rigid 3D transformation estimations, one for each vertebra $V_{i}, i\in\{1,2,\dots,5\}$ and incoming RGB-D frame $f$: $\hat{\textbf{T}}_{V_{i}}(f)$, $\hat{\textbf{T}}$ denoting that the transformation is estimated. The navigation module (Section \ref{reg-sec:navigation}) is responsible for tracking a surgical drill sleeve (rigidly attached to the drill) and finding its 3D transformation in each frame: $\textbf{T}_{D}(f)$. For each frame, the navigation module streams the aforementioned six transformations $(\hat{\textbf{T}}_{V_{i}}(f), i\in\{1,\dots,5\} \text{ and } \textbf{T}_{D}(f))$ to the HMD via a UDP connection. The second part of the navigation module consists of AR guidance for pedicle screw placement on the HMD, based on the received vertebra and drill sleeve poses. Finding the relative transformation ${}^{HMD}\textbf{T}_{S}$ between the coordinate frame of the RGB-D sensor ${S}$ and the HMD device is required upon re-positioning of the RGB-D sensor and is based on standard chessboard detection (\cite{opencv_library}).

The registration module and the first part of the navigation module were implemented as a real-time capable C++ application with an OpenGL (\cite{woo1999opengl}) window for live visualization and controlling purposes, referred to as the server app. Libraries and implementation details are provided throughout the following sections. The second part of the navigation module, the AR guidance for pedicle screw placement, was implemented in Unity (2019.4.39f1, Unity Technologies, San Francisco, CA, USA) and will hereafter be referred to as the client app.

\subsection{Registration module}
\label{reg-sec:registration}
The registration module has two goals: first, given an unoccluded RGB-D frame, referred to as the \textit{initial frame}, it estimates the 3D pose of each lumbar vertebra L1--L5 with respect to the sensor cordinate frame. Second, given subsequent frames of the same viewpoint with surgeon interaction, referred to as the  \textit{interaction frames}, it updates the poses of L1--L5, if visible. In our workflow, the surgeon positions the RGB-D sensor above the incision without any occlusion by personnel or instruments and initiates the process.

\begin{figure}[htb!]
\centering
\includegraphics[width=\textwidth]{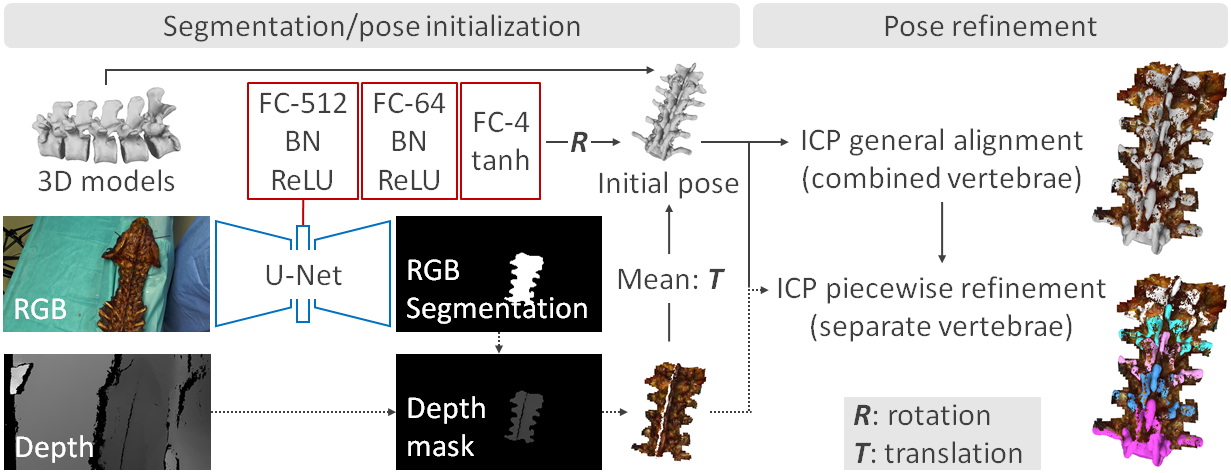}
\caption{For an \textit{inital frame}, all arrows are executed once: the network outputs result in a segmented point cloud and an initial pose ($R,T$) for the preoperative 3D models, which are then registered to the segmented point cloud (general alignment), followed by an individual registration (piecewise refinement). For subsequent \textit{interaction frames}, only the dotted arrows are executed, updating the poses of visible vertebrae individually.}
\label{reg-fig:overview}
\end{figure}

The registration method is illustrated in Fig. \ref{reg-fig:overview}. It is divided into segmentation/pose initialization and pose refinement. The first stage relies on a deep neural network that combines the concepts of 2D U-Net (\cite{ronneberger2015u}) and regression-based orientation prediction (\cite{mahendran20173d}). For a given \textit{initial frame} and during the inference time, the network outputs a 2D binary segmentation mask for the lumbar spine (segmentation path) and an estimate of the spine's rotation in the a system (orientation path) represented in form of a quaternion ($\textbf{R}$ in Fig. \ref{reg-fig:overview}). The segmentation output is used to mask the corresponding depth image, leading to a segmented point cloud of the lumbar spine. The preoperative 3D models are transformed with the predicted orientation as the rotation and the center of mass of the segmented point cloud as the translation ($\textbf{T}$ in Fig. \ref{reg-fig:overview}). In the second stage, an en-bloc registration of the combined preoperative 3D models is performed using ICP (\cite{Besl1992}) registration (general alignment), followed by a piecewise  ICP registration of each vertebra (piecewise refinement). Using the accurate pose determined from an \textit{initial frame}, efficient motion compensation in subsequent \textit{interaction frames} is achieved by iterative application of segmentation and piecewise refinement steps, indicated as dotted arrows in Fig. \ref{reg-fig:overview}.

\subsubsection{Data preparation}
\label{reg-sec:data_prepareation}
The SpineDepth dataset that was created in our preceding publication (\cite{liebmann2021spinedepth}) provides pose-annotated RGB-D recordings of mockup spine surgeries performed on ten cadaveric specimens. The extent of anatomical exposure in Specimen 1 was significantly less than in the other nine. It was therefore excluded from this study. Furthermore, we excluded Specimen 10 as its anatomy is extremely different from the remaining eight, i.e. it is much smaller. When the dataset was created, in each surgery, ten pedicle screws were placed bilaterally into vertebrae L1--L5. The placement of each screw was divided into four surgical steps, each captured in a separate recording by two downward facing RGB-D sensors simultaneously. After each screw placement, the sensors were repositioned to capture the surgical site from a new perspective. This resulted in a total of  80 recordings within the SpineDepth dataset from 20 different viewpoints per specimen. Within the same dataset and for each specimen, preoperative 3D models of vertebrae L1--L5, referred to as PreOp models, are available that are spatially aligned to the actual position of the anatomy. In other words, the SpineDepth dataset includes the aforementioned transformation $\hat{\textbf{T}}_{V_{i}}$ between each vertebra and the RGB-D frame observing it. However, in the dataset it is the ground truth transformation and therefore referred to as  $\textbf{T}_{V_{i}}$. Applying it on the PreOp 3D models transforms them to their ground truth location in the camera coordinate system of the respective RGB-D sensor.

In order to investigate the generalizability of our method to unseen anatomy, the data was prepared to support a leave-one-out cross-validation strategy with eight folds, one for each specimen. For each screw and surgical step three frames were selected, the first frame (\textit{initial frame}) and two random frames (\textit{interaction frames}), resulting in twelve frames per screw. The sensor viewpoints did not change within the four recordings. The \textit{initial frame} never contained surgeon interaction, while the \textit{interaction frames} had a chance thereof. The resulting 240 frames (10 screws $\times$ 12 frames $\times$ 2 RGB-D sensors) per specimen were used for training the network. They are referred to as \textit{training folds}. For later testing of the entire registration module, only the first two surgical steps, i.e. recordings, for each screw were considered, as they represent the relevant steps for navigation, i.e. entry point preparation and trajectory drilling with a surgical awl. These 40 recordings of a specimen were used in their full lengths with the first frame as \textit{initial frame} and all subsequent frames as \textit{interaction frames}. They are referred to as \textit{testing folds}.

The pose annotations provided by the SpineDepth dataset cannot be used directly for training our network, as the segmentation and orientation path of our network require a binary segmentation mask and a quaternion as ground truth, respectively. The former was generated as follows. A depth image of vertebrae L1--L5, transformed according to their ground truth poses, was rendered, using the method of \cite{Guney2015CVPR} and \cite{geiger2015joint}, and subtracted from the corresponding sensor depth image for all non-zero pixels in Matlab (R2021a, MathWorks, Portola Valley, CA, USA). The resulting mask consisted of all pixels with an absolute difference below 10 mm. This threshold correctly handled clear occlusions, while maintaining contiguous mask regions, despite measurement noise. Further smoothing of those regions was achieved by applying a 2D convolution with a kernel size of 15 and uniform weights of $1/225$, followed by a thresholding at 0.5, resulting in a binary mask again. The network orientation path only predicts the overall lumbar spine orientation (Fig. \ref{reg-fig:overview}), and not the one of each vertebra. Therefore, the rotation of vertebra L3 was stored as a quaternion for each frame, representing the overall spine orientation.

\subsubsection{Network architecture and training}
\label{reg-sec:network_arch_and_train}
The network architecture is indicated in Fig. \ref{reg-fig:overview}. The main structure is inspired  by U-Net, taking downsampled RGB images of size $144\times256$ ($H\times W$) as input. It consists of four downsampling blocks, each of the form \textit{Conv--BN--ReLU--Conv--BN--ReLU--MaxPooling}, with 64, 128, 256 and 512 filters of size $3\times3$. The bottleneck block is of the same form, with 512 filters, but without the \textit{MaxPooling} layer. The upsampling blocks mirror the downsampling ones, but with a leading upconvolution layer instead of a \textit{MaxPooling} layer at the end. Skip connections connect the down- and upsampling blocks. The output \textit{Conv} layer has 1 filter and \textit{sigmoid} activation, for which a dice loss $L_{D}$ is minimized. An additional branch is appended to the bottleneck where the $9\times16\times512$ feature representation is used for regression-based orientation prediction. After flattening, two blocks of \textit{FC--BN--ReLU}, with 512 and 64 units, follow. Another \textit{FC} layer with 4 units then predicts the spine orientation. The quaternion codomain of $[-1,1]$ is accounted for with a $tanh$ activation. For frame i, the geodesic loss $L_{G_i}$ is computed between the ground truth quaternion ${q_t}_i$ and its normalized predicted counterpart $q_{p_i}$, as given in Equation \eqref{eq:geodesic_loss}. Quaternions should have magnitude 1 in order to represent valid rotations. As suggested in \cite{langlois20183d}, a penalization term $L_{N_i}$ helped the network predicting such.

\begin{equation}
\label{eq:geodesic_loss}
    L_{G_i}=2\cos^{-1}{\lvert \langle {q_t}_i,q_{p_i} \rangle \rvert}
    \quad \textrm{and} \quad
    L_{N_i}=\left(1-\lvert\lvert q_{p_i} \rvert\rvert\right)^2
\end{equation}

\noindent Including $L_{D}$, the network's total loss for a batch of size $B$ is

\begin{equation}
    \mathcal{L}=\frac{1}{B}\sum_{i=1}^{B}L_{D_i}+\frac{1}{B}\sum_{i=1}^{B}\left(L_{G_i}+L_{N_i}\right).
\end{equation}

\noindent Augmentations were employed to enrich the SpineDepth dataset that included a limited number of specimens and viewpoints. 2D image rotation of $\alpha$ radians was applied on the input images and the corresponding ground truth quaternion by multiplication with quaternion $[\cos{\frac{\alpha}{2}}, 0, 0, \sin{\frac{\alpha}{2}}]$, which represents a rotation around the camera's z-axis. Each frame was augmented by rotations of 30, 90, 150, 210, 270 and 330 degrees, resulting in 1680 frames ($240+6\times240$) per \textit{training fold}. The 240 frames are: 10 screws $\times$ 12 frames $\times$ 2 RGB-D sensors (Section \ref{reg-sec:data_prepareation}).

The resulting network with $\sim$58M trainable parameters was implemented in Keras (2.7.0, \cite{chollet2015keras}). For each specimen, it was trained from scratch on the remaining eight \textit{training folds} during 30 epochs with batch size of 32 using the Adam optimizer (\cite{kingma2014adam}). The learning rate was set to $10^{-4-\lfloor\frac{epoch}{10}\rfloor}$. Training took roughly 30 minutes on a NVIDIA Quadro P6000. Three trainings were performed per specimen. For each specimen, the best results in terms of $L_{D}$ are presented.

\subsubsection{Registration and pose update}
\label{reg-sec:registration_and_pose_update}
\begin{figure}[htb!]
\centering
\includegraphics[width=\textwidth]{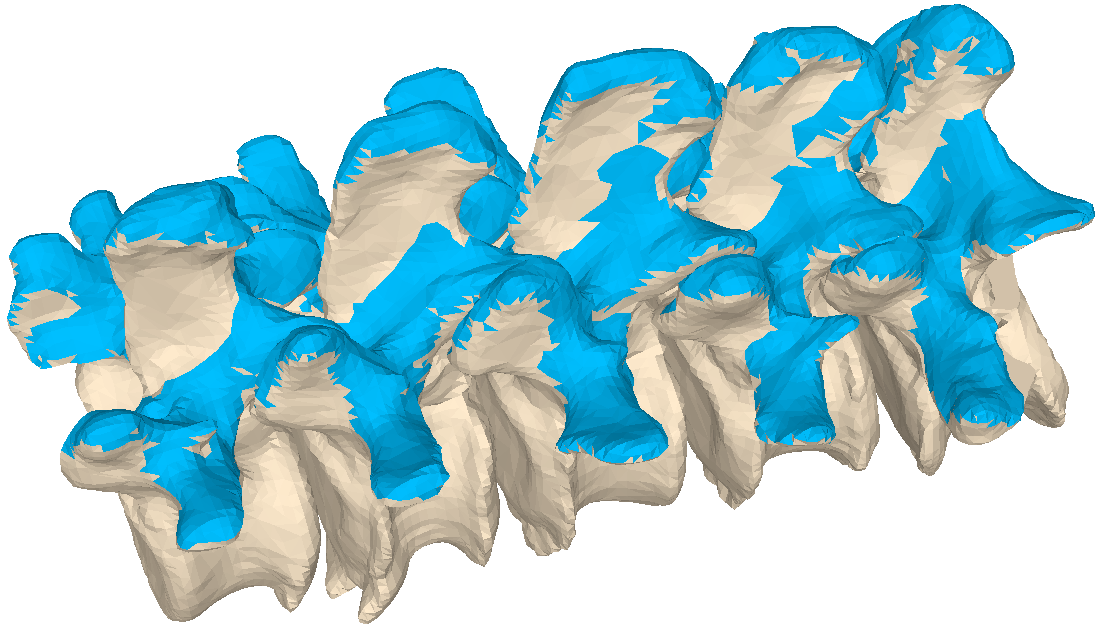}
\caption{Points used for registration and pose updates (blue).}
\label{reg-fig:selected_points}
\end{figure}

The registration for an \textit{initial frame} consists of an initial pose, a general alignment and a piecewise refinement (Fig. \ref{reg-fig:overview}). They are denoted as $\hat{\textbf{T}}_{init}$, $\hat{\textbf{T}}_{gen}$ and $\hat{\textbf{T}}_{ref_{V_{i}}}$. The following sections elaborate on each part followed by an explanation of the pose update for the \textit{interaction frames}. Note that only the points visible from an orthogonal posterior view were selected from the PreOp models (3-matic, Materialise NV, Leuven, Belgium) and used for registration and pose updates (Fig.\ref{reg-fig:selected_points}).

\paragraph{Initial pose estimation}
For an \textit{initial frame's} RGB image, our network predicts a resized binary segmentation mask $M_{p}$ ($1080\times1920$ pixels) and a normalized quaternion $q_{p}$. Using $M_{p}$, the corresponding full point cloud $PC_{f}$ is masked to produce a segmented point cloud $PC_{s}$. The initial pose for the first frame, which is used as the \textit{initial frame}, $\hat{\textbf{T}}_{init}$ for the PreOp models is constructed from the center of mass of the largest connected component (\cite{opencv_library}) of $PC_{s}$ of size $N$ as the translation and the inferred $q_{p}$ as the rotation:

\begin{equation}
    \hat{\textbf{T}}_{init} = [R, T] = \left[ q_{p}, \frac{1}{N}\sum_{n=1}^{N}{PC}_{s_n} \right]
\end{equation}

The trained network was converted to Open Neural Network Exchange (ONNX, \cite{onnx}) format and integrated into the server app using TensorRT (8.0.3, \cite{tensorrt}). This way, the RGB-D sensor data, which is available on GPU memory, can be directly fed to the network. Furthermore, $PC_{f}$ is masked by $M_{p}$ on the GPU using CUDA (8.0.3.4, \cite{cuda}), resulting in $PC_{s}$, consisting of $N$ points.

\paragraph{General alignment}
The general alignment consists of point-to-point ICP between the combined PreOp models of L1--L5, transformed by $\hat{\textbf{T}}_{init}$, and $PC_{s}$. It is denoted as $\textbf{T}_{\text{ICP}_{combined}}$. The pose after general alignment is:

\begin{equation}
    \hat{\textbf{T}}_{gen} = \hat{\textbf{T}}_{init}\textbf{T}_{\text{ICP}_{combined}}
\end{equation}

The Point Cloud Library (1.12.0, \cite{Rusu_ICRA2011_PCL}) implementation with a maximum correspondence distance of 5 mm and stopping criteria of 50 iterations or a transformation epsilon of $<10^{-8}$ was used.

\paragraph{Piecewise refinement}
The piecewise refinement is based on the general alignment result. In essence, this is another point-to-point ICP alignment. However, given the piecewise nature of this alignment, PreOp models L1--L5 are aligned individually. First, the nearest neighbor points for each PreOp model are found in the masked point cloud $PC_{s}$. Point correspondences with a Euclidean distance below 2.0 mm were considered inliers. The Umeyama method (\cite{umeyama1991least,eigenweb}) method was then used to find the optimal transformation, which was applied to the respective PreOp model. The two steps are repeated 50 times, or until the average distance between point pairs did not decrease anymore ($\epsilon=10^{-8}$). The ICP result for vertebra $i$ is denoted as $\textbf{T}_{\text{ICP}_{piecewise_{i}}}$. The pose after piecewise refinement is:

\begin{equation}
    \hat{\textbf{T}}_{ref_{V_{i}}} = \hat{\textbf{T}}_{init}\textbf{T}_{\text{ICP}_{combined}}\textbf{T}_{\text{ICP}_{piecewise_{i}}}\quad i \in\{1,\dots,5\}
\end{equation}

As the same functionality is employed for real-time pose updates during interaction frames (see next paragraph), the nearest neighbor search is performed in parallel on the GPU using CUDA. The EIGEN (\cite{eigenweb}) implementation of the Umeyama method is computationally inexpensive, and was therefore performed serially for each PreOp model on the CPU.

\paragraph{Pose update}
After performing the registration based on the \textit{initial frame}, the poses of PreOp models L1--L5 are updated individually once a new frame is available. After network inference on the new frame, the same technique as for the piecewise refinement is used. Given that the extent of change in the vertebra poses is minimal for a single patient during an intervention and in hopes of boosting the performance, only a single iteration was performed, denoted as $\hat{\textbf{T}}_{update_{i}}(f)$ for vertebra $i$ in frame $f$. Furthermore, due to surgeon interactions, the visibility of the anatomy during \textit{interaction frames} may be obscured. Consequently, only PreOp models that have at least 90\% of the number of inliers as after the last iteration during piecewise refinement are updated. The pose $\hat{\textbf{T}}_{V_{i}}(f)$ (Fig. \ref{reg-fig:setup}) in any \textit{interaction frame} $f$ after transformation of each vertebra by $\hat{\textbf{T}}_{ref_{V_{i}}}$ is:

\begin{equation}
    \hat{\textbf{T}}_{V_{i}}(f) = \hat{\textbf{T}}_{V_{i}}(f-1)\textbf{T}_{upd_{i}}(f) \quad i \in\{1,\dots,5\}, f \in\{2,3,\dots\}
\end{equation}

\subsection{Navigation module}
\label{reg-sec:navigation}
The navigation module comprises two parts: tracking of a surgical drill sleeve and AR guidance for pedicle screw placement on the HMD. Both parts are explained in the next sections.

\subsubsection{Drill sleeve tracking}
\begin{figure}[htb!]
\centering
\includegraphics[width=\textwidth]{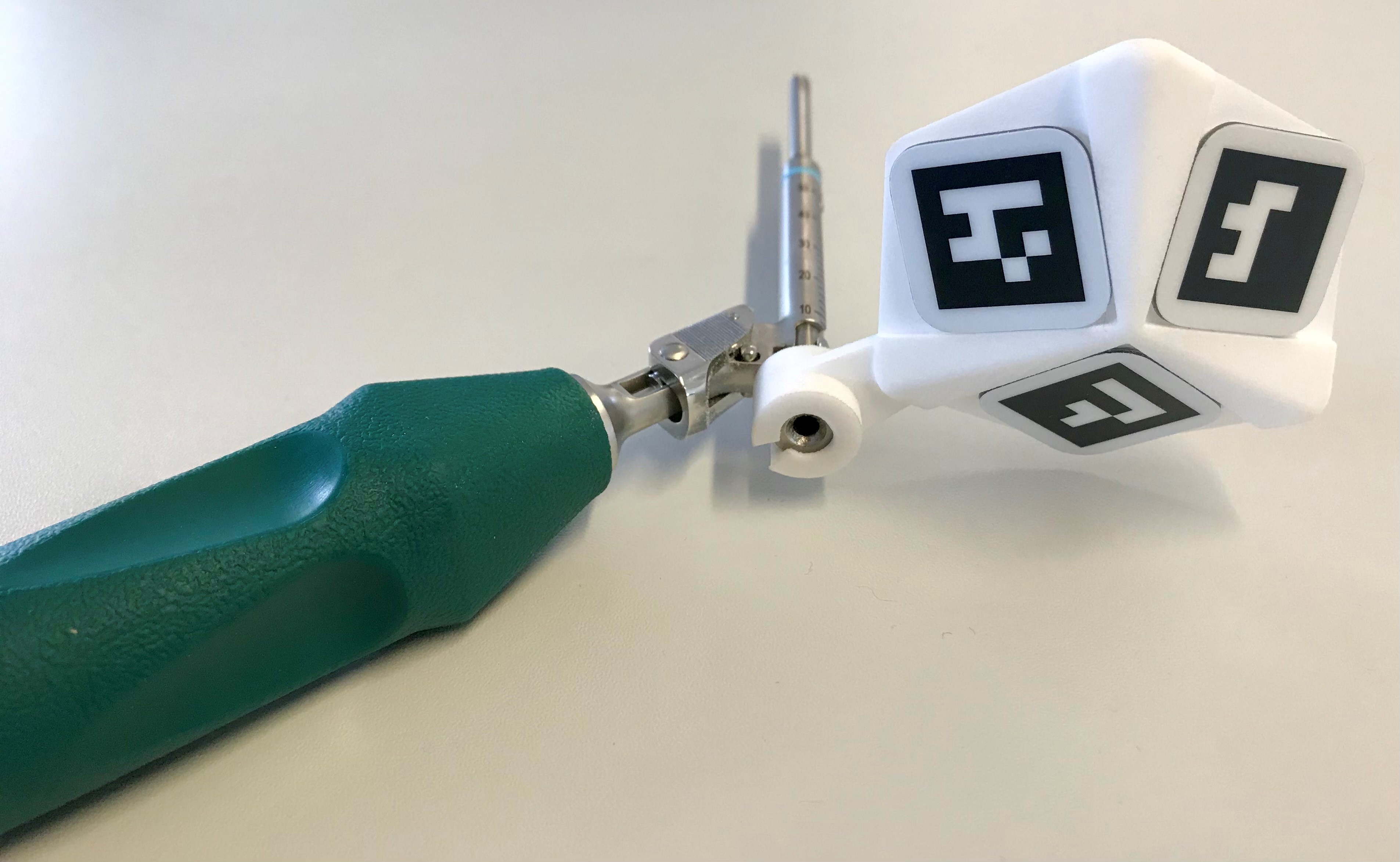}
\caption{Drill sleeve with three nonplanar, sterile markers for tracking.}
\label{reg-fig:sleeve}
\end{figure}

Tracking of the surgical drill sleeve was based on a custom-made, 3D printed component attached to the drill sleeve (Fig. \ref{reg-fig:sleeve}). The component was equipped with three nonplanar, sterile markers (Clear Guide Medical, Baltimore MD, USA) showing unique AprilTag (\cite{olson2011apriltag,wang2016apriltag}) patterns. The tracking was integrated into our server app and was performed on a separate thread, which is initiated upon app startup. Whenever a new frame was available from the RGB-D sensor, the undistorted left and right grayscale images were made available to the tracking thread. In both images, the markers were detected by the ArUco (\cite{MUNOZSALINAS2018158,Garrido2014,Garrido2016}) library. If at least two corresponding markers were found in both images, the respective 2D corner coordinates are used for triangulation (\cite{opencv_library}), yielding eight or twelve 3D corner coordinates. The actual pose is found by applying the Umeyama method between the ground truth corner coordinates, which are known by design, and their estimated counterparts resulting from the triangulation. A Kalman filter (\cite{Kalman1960,opencv_library}) with a constant acceleration model is used for noise reduction on the final drill sleeve pose.

\subsubsection{AR guidance}
\label{reg-sec:ar_guidance}
The goal of the AR guidance for pedicle screw placement is the accurate and fast navigation of the screw entry point and trajectory. Upon startup of the client app, the HMD establishes a UDP connection to the server app and continuously receives the poses of PreOp models L1--L5 as well as the drill sleeve pose. The surgeon positions the RGB-D sensor such that a reasonable initial pose, which is visualized on a monitor in the periphery of the OR, is estimated by the server app. A standard chessboard (\cite{opencv_library}) is used to co-calibrate the coordinate frame of the RGB-D sensor and the one of the HMD. The surgeon can trigger the detection in the client app by speech command. As soon as the chessboard is removed, the surgeon initiates the registration on an \textit{initial frame}, which is followed by pose updates in subsequent \textit{interaction frames} (Section \ref{reg-sec:registration_and_pose_update}).

\begin{figure}[htb!]
\centering
\includegraphics[width=\textwidth]{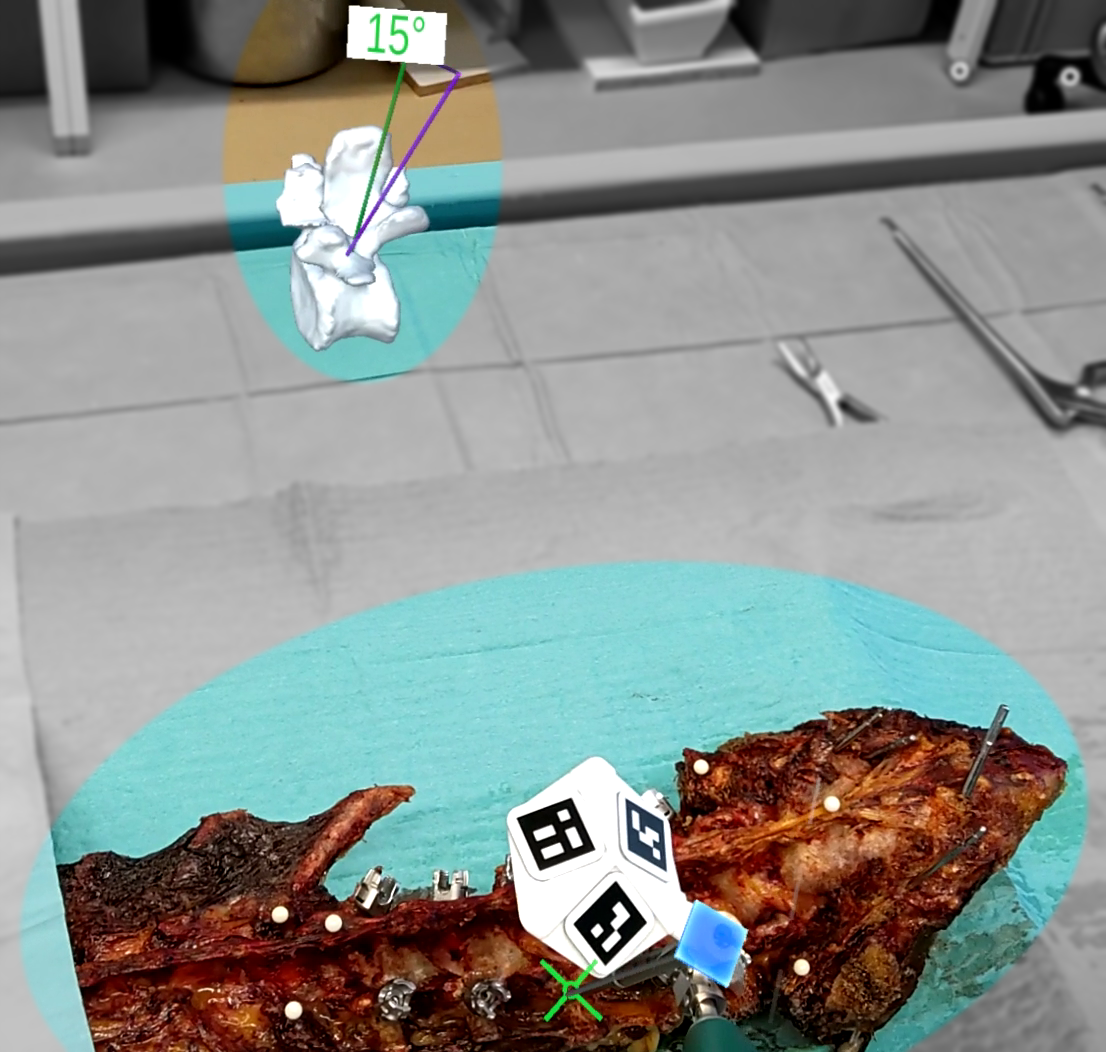}
\caption{Surgeon's view of AR navigation. The PreOp model is rendered as an axis-aligned virtual twin (top) and the current drill sleeve pose is visualized with respect to the preoperatively planned screw entry point and trajectory. In addition, the angular 3D deviation between the drill sleeve and the screw trajectory is shown. The direct overlay of the entry point (green cross) can be shown/hidden by the surgeon. Note that white push-pins are only used for postoperative evaluation of target registration error.}
\label{reg-fig:navigation}
\end{figure}

In the client app, the surgeon is provided with three different visualization components. The most important component is the virtual twin presented in the work of \cite{wolf2023different}, who investigated different user interfaces for AR-guided pedicle screw placement. Their virtual twin approach, where PreOp models and navigation information are not directly overlaid onto the anatomy, but rendered in an axis-aligned fashion with only a translational offset from the anatomy (Fig. \ref{reg-fig:navigation}), was integrated into the client app. \cite{wolf2023different} showed that this approach allows for accurate screw placement, while both ease of use and cognitive load were well rated by surgeons. On the virtual twin, the current drill sleeve pose was visualized with respect to the preoperatively planned screw entry point and trajectory. In addition, the angular 3D deviation between the drill sleeve and the screw trajectory was shown. Besides the virtual twin, a direct overlay of the entry point in form of an aiming cross could be shown/hidden by the surgeon. Lastly, the PreOp models could also be visualized on the anatomy upon request. This was particularly useful to qualitatively check the overall registration accuracy. For a more detailed verification of the registration, the surgeon was asked to touch certain anatomical landmarks using the drill sleeve on the anatomy and confirm their correspondence on the virtual twin before starting the actual navigation of a screw. If the registration was unsatisfactory, the RGB-D sensor was repositioned and the process was repeated from the co-calibration on. After successful navigation of a level, the surgeon selects the next level in the client app menu.

\subsection{Evaluation}
\label{reg-sec:evaluation}
The method was evaluated in two stages. First, the registration module was evaluated separately on the SpineDepth dataset, referred to as the verification. The entire prototype (registration and navigation module) was then evaluated in a cadaveric experiment on an unseen lumbar spine anatomy, where ten real pedicle screws were placed under AR guidance, referred to as \textit{ex-vivo} validation. The two evaluation stages and the respective outcome measures are described in the following sections.

\subsubsection{Verification}
For verification, the eight trained networks (Section \ref{reg-sec:network_arch_and_train}) were employed. For each specimen, each of the 40 recordings (Section \ref{reg-sec:data_prepareation}) in the respective \textit{testing fold} was evaluated using the server app, with the first frame as the \textit{initial frame} and all subsequent frames as \textit{interaction frames}.

The recordings in the SpineDepth dataset were made from a broad variety of viewpoints, some of which providing a strongly inclined lateral view of the anatomy, thus potentially influencing the registration quality. Preliminary analysis had shown that the 3D angle between the RGB-D sensor's forward axis and the coronal plane normal correlate positively with the target registration error (TRE) after applying our proposed method. The correlations (Pearson correlation coefficient: PCC) are reported for each specimen in the \textit{testing fold} (viewpoint-error correlation, VpErCo). It was defined that only recordings with a 3D angle below 30° are considered. The number of recordings fulfilling this criterion is also part of the results (acceptable viewpoints, AcVp). Note that for generalization purposes, these viewpoints were not excluded from network training.

A single threshold for the accuracy of pedicle screw placement with respect to the  optimally planned screw could not be defined, as the required accuracy is generally dependent on different anatomical and surgical factors such as the anatomical morphology and pathology of the patient, the underlying bone quality, or the utilized surgical approach. Furthermore, with an automated evaluation based on the given dataset, tapping landmarks on the anatomy to confirm the registration accuracy, as done during AR guidance (Section \ref{reg-sec:ar_guidance}), is not possible. Therefore, a successful registration was measured by the established clinical criteria according to \cite{Modi2008}, who define a screw perforation of less than 2 mm as safe. To this end, optimal pedicle screws (\diameter: 5 mm) were planned bilaterally using an in-house developed preoperative planning software (CASPA, University Hospital Balgrist, Zurich, Switzerland). For the assessment of pedicle perforation, a 3D model of the pedicle was extracted from the PreOp model and imported into MATLAB. The screws were represented as cylinders (\diameter: 5 mm). For each frame of a recording, the screws were transformed according to the corresponding vertebra pose found by our method, while the pedicle 3D model was transformed according to the respective ground truth pose. For all points of the pedicle 3D model, it was verified whether they are located inside the cylinder representing the pedicle screw. If no points were inside the cylinder, there was no perforation. Otherwise the perforation was quantified as the maximum distance from all points inside the cylinder to the cylinder surface. Note that the registration success was defined on a per frame basis and for the target screw only (the screw that the surgeon works on in the respective recording), e.g. if the surgeon prepares the entry point of L2 left in the recording, the registration for a frame was considered successful when the previously described perforation assessment using the estimated pose for L2 ($\hat{\textbf{T}}_{V_{2}}$) revealed that the perforation would have been below 2 mm. The success rate of a single recording was defined as the number of successful frames divided by the total number of frames. The success rate of an entire specimen equals the average success rate over all recordings in the \textit{testing fold}. In the same way, the average 3D angular deviation between the optimal and estimated screw trajectory (trajectory error: $E_{E_TR}$) as well as the average 3D distance between the optimal and estimated screw entry point (entry point error: $E_{E_EP}$) are reported, except that for an entire fold the median over all 40 recordings is reported. As the TRE considers the registration for an entire vertebra, it can only be computed on a per vertebra level. The average TRE for exemplary vertebra L2 in a recording of $F$ frames with $K=3$ landmarks ($L_{1}$: spinous process, $L_{2}$ and $L_{3}$: left and right transverse processes) and $\text{d}(p_{1},p_{2})$ as the 3D Euclidean distance between two points $p_{1}$ and $p_{2}$ is defined in Equation \eqref{eq:tre}. Again, the median over all 40 recordings is reported.

\begin{equation}
\label{eq:tre}
    \text{TRE} = \frac{1}{FK}\sum_{f=1}^{F}\sum_{k=1}^{K}\text{d}(\textbf{T}_{V_{2}}(f){L_{k}}, \hat{\textbf{T}}_{V_{2}}(f)L_{k}).
\end{equation}

Preliminary analysis showed that the alignment of each PreOp model can improve during the first few \textit{interaction frames}, due to the slightly varying 3D reconstructions provided by the RGB-D sensor. Therefore, the TRE is reported as of frame 61 ($\sim 2$ s).

As an additional result, the percentage of updated poses ($\%_{update}$) for the vertebra of interest in \textit{interaction frames}, i.e. the number of actual pose updates according to our method (Section \ref{reg-sec:registration_and_pose_update}) divided by the number of possible frames are reported.

Besides the outcome measures related to registration and pose updates, the performance of the eight trained networks are reported. Segmentation accuracy was evaluated with the Dice similarity coefficient (DSC). As in \cite{tulsiani2015viewpoints} and \cite{mahendran20173d}, the orientation prediction was evaluated with the median geodesic angle error (MGAE), which equals the median loss defined in Equation \eqref{eq:geodesic_loss} over an entire fold, expressed in degrees. Note that these outcome measures are based on number of frames defined for the \textit{training folds}, i.e. not full recordings but 240 frames per fold/specimen.

\subsubsection{Ex-vivo validation}
The goal of the ex-vivo validation was to place ten pedicle screws (L1--L5, left and right) under AR guidance using the herein presented method (registration and navigation module) on an unseen lumbar spine. A fresh frozen specimen was used. Ethical approval was obtained from the ethical committee of Canton Zurich (Basec-Nr. 2017-00874). The specimen was CT scanned using a NAEOTOM Alpha\textsuperscript{\textcopyright} device (Siemens Healthineers, Erlangen, Germany) with a 0.8 mm slice thickness and a 0.41$\times$0.41 mm in-plane resolution (x-y). 3D models of L1--L5 were extracted using the global thresholding, region growing and wrapping functionalities of the Mimics software (Materialise NV, Leuven, Belgium). Again, the points visible from an orthogonal posterior view were selected as described in Section \ref{reg-sec:registration_and_pose_update} (Fig. \ref{reg-fig:selected_points}). Optimal pedicle screws (\diameter: 5 mm) were planned in CASPA. In preparation of the experiment, the specimen was thawed and dissected to have no soft tissues, e.g. paravertebral muscles, without damaging the intraspinous ligament, the ligamentum flavum as well as the facet joint capsule. The specimen was fixated to a wooden board with surgical pins through spinal levels T6/7 and S1.

The network was trained in the same way as described in Section \ref{reg-sec:network_arch_and_train}, but using the \textit{training folds} of all eight specimens in the SpineDepth dataset. As there was no ground truth available for the unseen specimen, the number of epochs was reduced to ten to mitigate overfitting to the experimental setup, e.g. the spine orientation w.r.t. the table, of the SpineDepth dataset. Due to the fact that the preoperative CT was taken from the frozen specimen, the inter-vertebral deformation between the pre- and intraoperative states was higher than in the SpineDepth dataset, where the CT was conducted in a fully thawed state. Therefore, the piecewise refinement (Section \ref{reg-sec:registration_and_pose_update}) was performed for 50 iterations, without a stopping criterion, to overcome local minima due to the 2 mm inlier threshold.

\begin{figure}[htb!]
\centering
\includegraphics[width=10cm]{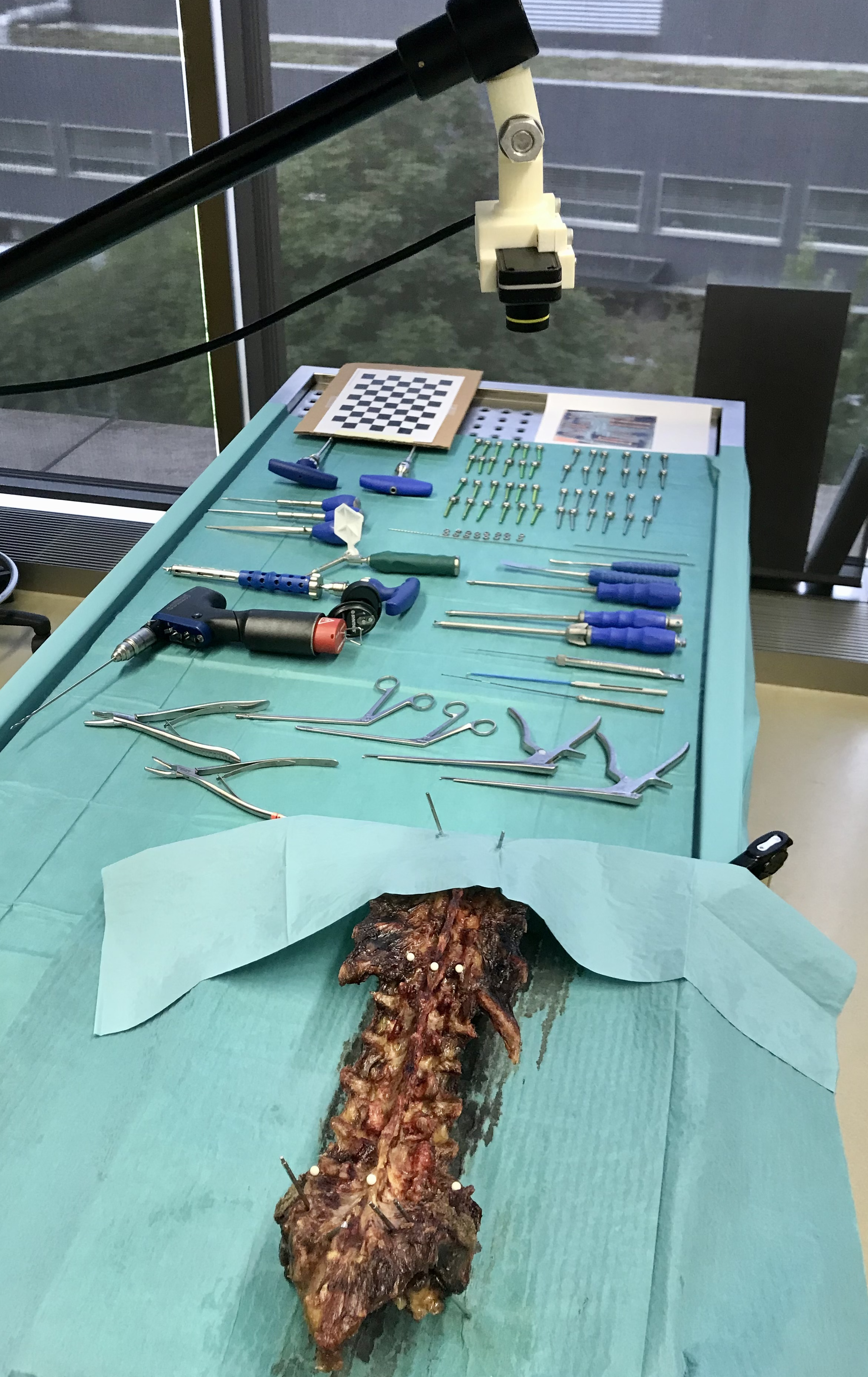}
\caption{Setup during the ex-vivo validation. After inserting the five pedicle screws on the right side, the specimen was rotated by 180°. After re-registration, the surgeon could insert the five screws on the left side.}
\label{reg-fig:exp_setup}
\end{figure}

During the experiment, the RGB-D sensor was placed above the surgical site (Fig. \ref{reg-fig:exp_setup}) and the server and client apps were started. After that, the workflow was as described in Section \ref{reg-sec:ar_guidance}: the sensor viewpoint was adjusted, such that the initial pose was reasonable, followed by the co-calibration of sensor and HMD. The registration and pose updates were initiated. For each vertebra, the surgeon checked the registration accuracy and inserted the respective pedicle screw (right side) according to the AR guidance. For screw insertion on the left side, the specimen was rotated 180° (the other side of the table was not optimal to stand for the surgeon), followed by a re-registration.

For the ex-vivo validation, TrEr, EpEr and TRE are reported. TrEr and EpEr were quantified following the same procedure as described in \cite{liebmann2019pedicle}. A postoperative CT of the specimen was acquired with the same imaging device and protocol as for the preoperative scans. 3D models of the bone anatomy and the screws were extracted. In the CASPA software, the PreOp models along with the planned screw trajectories were registered to the postoperative bone anatomy using point-to-plane ICP (\cite{rusinkiewicz2001efficient}). In the same fashion, generic cylindrical 3D models were aligned to the postoperative screw 3D models. The cylinders' main axes were compared to the planned screw trajectories, yielding the 3D angular deviation TrEr. The 3D Euclidean distance EpEr was determined by comparing the planned entry points to the intersection point of the cylinders' main axes with the registered preoperative 3D model.

In contrast to the SpineDepth verification where the ground truth vertebral poses were available, the data collected in the \textit{ex-vivo} validation experiment lacked the registration ground truth; therefore, the experiment was captured as an RGB-D recording and the TRE was quantified retrospectively in a static manner.  Six push-pins were inserted into the spinal levels T12 and S1 (three each) before the experiment (Fig. \ref{reg-fig:navigation}). The 3D positions of the push-pin head centers were determined in the postoperative CT using the Mimics software as well as in the left and right RGB images of the RGB-D \textit{initial frame} using blob detection and triangulation techniques (\cite{opencv_library}). The best fit in a least-squares sense between the two point sets was found in the CASPA software and allowed for transforming the preoperative 3D models into the coordinate frame of the RGB-D sensor. The TRE is based on the same three landmarks per vertebra as for the verification and is also reported for the 61\textsuperscript{st} frame ($\sim$ 2 s) after the \textit{initial frame}.

Additional outcome measures are: the registration time, the time for pose updates, and the navigation time. The latter was defined according to \cite{farshad2021operator}, as the time from picking up the drill sleeve until the drilling process was started.

\subsubsection{Ablation study}
To further understand the capabilities of the proposed registration method and the mechanisms leading to our results, an ablation study was conducted. For both verification stages, the server app was run three times with the following modifications (italic font denotes the name of the modification used hereinafter):
\begin{itemize}
    \item \textit{General:} Registration only included general alignment, no piecewise refinement, no pose updates
    \item \textit{Refinement:} Registration included general alignment and piecewise refinement, but no pose updates
    \item \textit{First-60:} Registration included general alignment, piecewise refinement and pose updates during the first 60 \textit{interaction frames} of a recording
\end{itemize}

For the ablation study, only the TRE is considered. Note that, for the ex-vivo validation, {First-60} is equal to our primary results by definition and is therefore not reported.

\section{Results}

Table \ref{reg-tab:results} summarizes the results. It comprises the verification and the ex-vivo validation.

\begin{table*}[htb!]
\caption{Results overview. Specimen numbering is according to the SpineDepth dataset. Spec.: Specimen. Perf.: Performance. Ex-vivo: Ex-vivo validation. VpErCo: viewpoint-error correlation. PCC: Pearson correlation coefficient. AcVp: acceptable viewpoints. SuRe: successful registrations. TrEr: trajectory error. EpEr: entry point error. TRE: target registration error. UpPo: updated poses. DSC: Dice similarity coefficient. MGAE: median geodesic angle error. N/A: not applicable.}
\label{reg-tab:results}
\centering
\begin{tabular}{c c c c c c c c c c}
\toprule
\multirow{2}[6]{*}{Spec.} & \multicolumn{7}{c}{Registration and pose updates} & \multicolumn{2}{c}{Network perf.} \\
\cmidrule(lr){2-8} \cmidrule(lr){9-10}
& VpErCo & AcVp    & SuRe & TrEr & EpEr & TRE  & UpPo & DSC & MGAE \\ 
& [PCC]  & (of 40) & [\%] & [°]  & [mm] & [mm] & [\%] &     & [°]  \\
\midrule
2&0.43&28&90&2.09&3.50&4.12&10&0.67&13\\
3&0.40&22&99&2.24&2.66&3.14&16&0.74&14\\
4&0.78&22&94&1.71&1.89&2.47&22&0.75&18\\
5&0.67&24&92&1.27&1.45&1.51&15&0.76&14\\
6&0.27&30&100&1.06&1.14&1.28&9&0.74&13\\
7&0.12&36&100&1.61&1.72&1.79&17&0.74&16\\
8&0.39&36&94&1.94&3.16&3.43&23&0.74&13\\
9&0.73&26&100&2.42&3.95&4.10&22&0.67&21\\
\midrule
Mean&0.47&28&96&1.79&2.43&2.73&17&0.72&15\\
SD&0.23&6&4&0.47&1.03&1.13&5&0.04&3\\
\midrule
Ex-vivo&N/A&N/A&N/A
&2.55
&1.95
&1.20
&N/A&N/A&N/A\\
SD&&&&1.67&1.07&0.22&&&\\
\bottomrule
\end{tabular}
\end{table*}

For the verification, the viewpoints and TRE correlated by $0.47\pm 0.23$. The number of acceptable viewpoints was $28\pm 6$, with an average of $96\pm 4$\% of registrations being successful. The median trajectory error was $1.79\pm 0.47$° and the median entry point error of $2.43\pm 1.03$ mm. The median TRE was $2.73\pm 1.13$ mm. An average of $17\pm 5$\% of poses were updated during \textit{interaction frames}. The mean DSC was $0.72\pm 0.04$ and the MGAE $15\pm3$°.

\begin{figure}[htb!]
\centering
\begin{subfigure}{0.3275\textwidth}
    \centering
    \includegraphics[width=\textwidth]{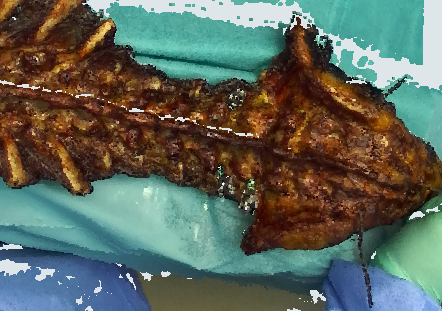}
    \caption{}
\end{subfigure}
\begin{subfigure}{0.3275\textwidth}
    \centering
    \includegraphics[width=\textwidth]{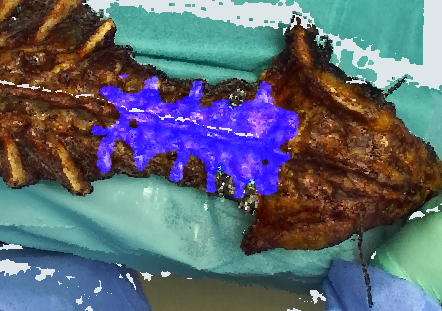}
    \caption{}
\end{subfigure}
\begin{subfigure}{0.3275\textwidth}
    \centering
    \includegraphics[width=\textwidth]{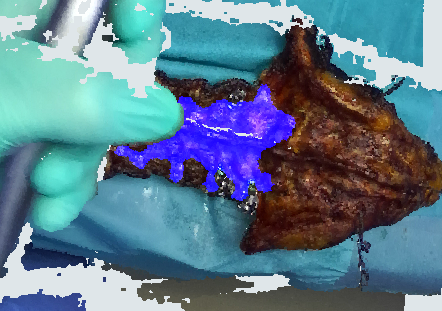}
    \caption{}
\end{subfigure}
\begin{subfigure}{0.3275\textwidth}
    \centering
    \includegraphics[width=\textwidth]{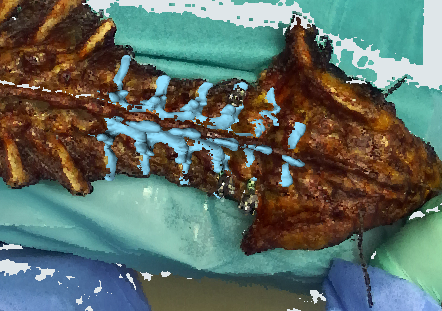}
    \caption{}
\end{subfigure}
\begin{subfigure}{0.3275\textwidth}
    \centering
    \includegraphics[width=\textwidth]{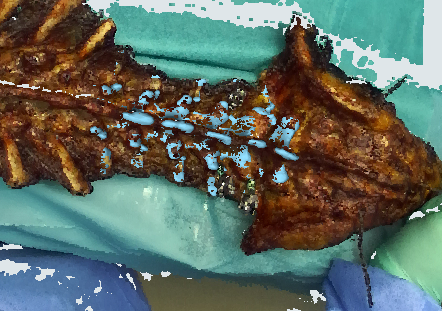}
    \caption{}
\end{subfigure}
\begin{subfigure}{0.3275\textwidth}
    \centering
    \includegraphics[width=\textwidth]{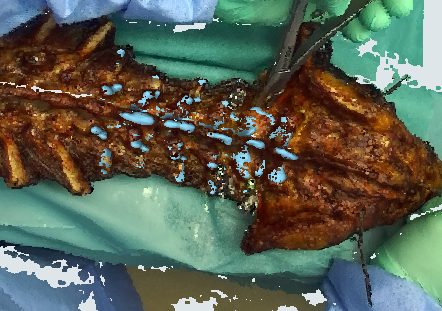}
    \caption{}
\end{subfigure}
\begin{subfigure}{0.3275\textwidth}
    \centering
    \includegraphics[width=\textwidth]{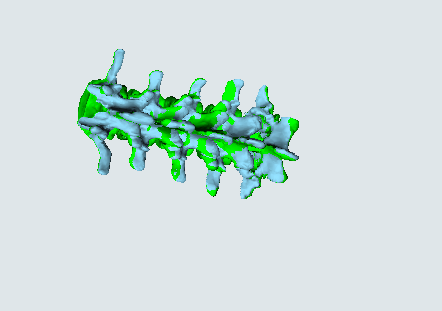}
    \caption{}
\end{subfigure}
\begin{subfigure}{0.3275\textwidth}
    \centering
    \includegraphics[width=\textwidth]{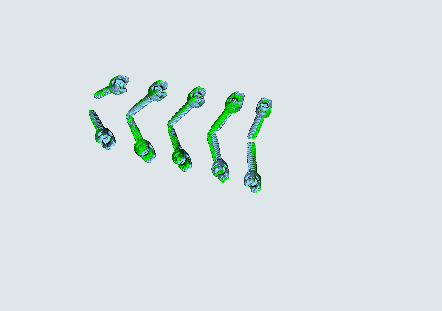}
    \caption{}
\end{subfigure}
\begin{subfigure}{0.3275\textwidth}
    \centering
    \includegraphics[width=\textwidth]{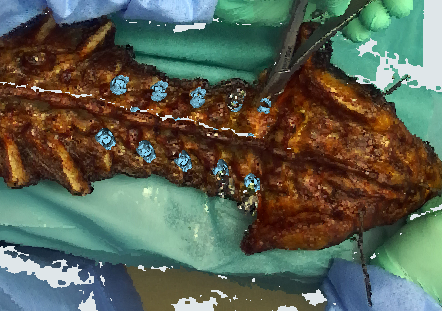}
    \caption{}
\end{subfigure}
\caption{Exemplary recording from verification. (a) Point cloud (b) Segmentation with screw occlusion handling (L4 left \& right, L5 left). (c) Segmentation with surgeon occlusion handling. (d) Initial pose in \textit{initial frame}. (e) General alignment in \textit{initial frame}. (f) After piecewise refinement and 60 pose updates. (g) Estimated (blue) and ground truth (green) vertebra poses. (h) Estimated (blue) and ground truth (green) simulated screws. (i) Estimated simulated screws on point cloud.}
\label{reg-fig:results}
\end{figure}

An exemplary case from the verification is illustrated and explained in Fig. \ref{reg-fig:results}. It shows segmentation and occlusion handling, initial pose, general alignment, piecewise refinement and pose updates as well as comparison of ground truth models and screws to their counterparts estimated by the proposed registration module.

During the ex-vivo validation, three registrations were necessary. After placing the first screw (L1, right), the client app crashed unexpectedly. Therefore a second registration became necessary. After placement of the remaining four screws on the right side, the specimen was rotated by 180°, followed by the third registration, such that the surgeon could operate on the left side. The mean trajectory error was 2.55$\pm$1.67°, while the mean entry point error was 1.95$\pm$1.07 mm. All screws were of grade 0, i.e. fully contained within the pedicle (\cite{Modi2008}). The mean TRE was $0.87\pm 0.31$ mm for the first, $0.94\pm 0.19$ mm for the second, and $1.80\pm 0.62$ mm for the third registration, respectively. The mean navigation time per screw was $28\pm 11$ s.

For the verification, the ablation study showed a median TRE of $2.66\pm 1.05$ mm for \textit{First-60}. This was slightly more accurate than \textit{Refinement} ($2.68\pm 1.14$ mm), which in turn was slightly more accurate than our primary results ($2.73\pm 1.13$ mm). The least accurate was \textit{General} with a median TRE of $2.89\pm 1.07$ mm.

For the ex-vivo validation, our primary results were the most accurate with a TRE of $1.20\pm 0.22$ mm, followed by \textit{Refinement} ($1.38\pm 0.45$ mm) and \textit{General} ($2.53\pm 0.23$ mm).

\section{Discussion}
Despite the fact that CAOS can increase accuracy as well as safety in complex orthopedic procedures, such as pedicle screw placement (\cite{gelalis2012accuracy,perdomo2019accuracy}), the clinical adoption of such methods is arguably low (\cite{joskowicz2016computer,hartl2013worldwide,nadeau2015qualitative}). Besides economic reasons, one major barrier along ubiquitous adaptation of the existing CAOS solutions is their interference with the standard surgical workflow. More specifically, main limiting factors associated with the current CAOS systems for surgical navigation can be noted as: cumbersome and time-consuming, ionizing radiation exposure, lengthy registration procedures and unintuitive visualization of spatial navigation information on 2D monitors in the OR periphery. In this work, we intended to tackle these drawbacks and presented a simplistic and radiation-free approach for automatic, accurate and fast pedicle screw placement in cadaveric lumbar spines under AR guidance.

The verification on the SpineDepth dataset showed a registration success rate of 96\%, meaning that the target screw would have been placed successfully within the clinical safe zone in 96\% of the frames. In the study of \cite{felix2021towards}, who pursued a similar approach for femur and tibia, the success of a registration was defined based on the percentage of inliers, which had to be at least 80. They reached a success rate of 37.7\% for the femur and 35.2\% for tibia, respectively. The required registration accuracy, defined as 3° rotational error and 3-4 mm translational error, was only met in terms of translation. The surface-based femur registration and tracking approach of \cite{hu2022automatic} achieved a root-mean-square error of 2.40 mm on real-time captures of a bone phantom, which reduced to 2.07 mm when the bone phantom data was processed with the suggested PointNet-based restoration network. For spine surgery, a wide range of acceptable registration errors can be found in the literature, which depend on various factors. \cite{rampersaud2001accuracy} defined that the maximum rotational and translational deviation for the lumbar spine reach from 2.1°/0.65 mm (L1) to 12°/3.8 mm (L5) for screws with a diameter of 6.5 mm. As the TRE reported in this work comprises the rotational and translational aspect, a comparison to TrEr and EpEr is more meaningful. While the TrEr (1.79°) is within the aforementioned limits in our case, the EpEr (2.43 mm) exceeds the limit for L1. Besides the targeted spine level, different methods for error calculation can affect the reported values (\cite{holly2007image}). The TRE is a well-known measure to characterize the accuracy of navigation approaches (\cite{ershad2014minimization}). \cite{guha2019intraoperative} investigated the error propagation of clinical-grade navigation systems w.r.t. a dynamic reference frame (DRF) attached to the anatomy, which is a common motion compensation technique, on four human cadavers. They compared intraoperative tip positions of a tracked awl (mimicking a bone screw) to typical pedicle screw entry points in a postoperative CT. An average 3D navigation (note that this error may difer from the registration error) error of 2.71 mm at DRF level was found. This error increased with a larger distance to the DRF level. Although the respective registration error must have been lower, the fact that the use of a DRF is the gold standard makes it eligible for comparison to our registration error, assuming tracking errors in current navigation systems are minimal: the EpEr (2.43 mm) of our verification was superior and the TRE (2.73) equal. When comparing to the TRE of $1.43 \pm 0.35$ mm in the semi-automatic microscopic RGB stereo method of \cite{ji2015patient}, three out of eight specimens in the verification can be considered within the range of their standard deviation. The required 2 mm maximum acceptable registration error for cranial and spine procedures (\cite{faraji2020machine}) is reached for three of our specimens. However, it should be considered that the dataset already comes with certain inaccuracies (ground truth TRE of 1.5 mm). The first (0.87 mm) and second (0.94 mm) registration in our ex-vivo validation showed sub-millimetric accuracy, which is equal or close to studies using navigation systems with manual point sampling for pedicle screw navigation (0.9 mm in \cite{papadopoulos2005accuracy}, 0.7 mm in \cite{nottmeier2007timing}) or cutting-edge intraoperative CT device for cranial procedures (0.93 mm in \cite{carl2018intraoperative}). The screw accuracies with a TrEr of 2.55° and EpEr of 1.95 mm in the ex-vivo validation are in line with other studies investigating surgical navigation for pedicle screw placement. The AR system used in \cite{felix2022augmented} achieved a 3D accuracy of 2.5° and 1.9 mm for open surgery in cadavers. In \cite{van2015clinical}, the accuracy of 178 minimally invasive screws using a robotic system was assessed, resulting in a mean 2D in-plane error of 2.55° and a 3D entry point deviation of 2.0 mm. An even lower mean angular deviation of 1.53° can be found in the cadaveric study of \cite{lamartina2015pedicle}. However, again, the values originate from 2D in-plane measurements.

Besides showing similar accuracy, our registration method has two advantages over clinically established navigation systems. First, the registration is fully automated and is performed for all targeted levels simultaneously, while  computation time required by our method was less than 2 s (and after that real-time) considerably lower to other clinical-grade systems based on surface data (less than 20 s in \cite{faraji2020machine}) or manual point sampling (117 s in \cite{nottmeier2007timing}, 125 s in \cite{farshad2021operator}). Our ablation study shows that the piecewise refinement improves accuracy especially when the preoperative images were acquired in a different patient positioning. Second, anatomy displacement induced by surgical manipulation or respiration can be as high as $1.85\pm 1.48$ mm and $1.09\pm 0.44$ mm, respectively (\cite{guha2019intraoperative}). Our ablation study could not show superior accuracy when applying real-time pose updates in \textit{interaction frames} throughout entire recordings (primary results) as opposed to a registration based on an \textit{initial frame} only (\textit{Refinement}) or applying updates for the first 60 \textit{interaction frames} (\textit{First-60}). One of the main reasons could be the frame rate of the RGB-D sensor as well as motion blur in the images, leading to a insufficient 3D reconstruction. The surgical interactions in the SpineDepth dataset are of fast nature. However, slower motions, such as breathing, could be compensated with the method at hand (upon proper investigations in the future). For faster motion, different sensor types, not based on RGB or grayscale images, could further improve the performance of our method in this regard. Finally, the anatomical part that is moved the most is (partly) hidden by the surgeon, and therefore challenging to track. Nevertheless, we see our approach being a foundation for developing automatic level-wise motion compensation in real-time without needing a DRF clamped to the anatomy.

The average time for pre-drilling a screw trajectory with  method was 28 s per screw which can be considered as very fast. This is superior to navigation using C- or O-arm (248 s for C-arm and 134 s for O-arm in \cite{liu2017comparison}), as well as other studies using AR guidance in cadaveric specimens (57.5 s in \cite{muller2020augmented}, 67 s in \cite{farshad2021operator}) or a first in-human study (312 s in \cite{elmi2019pedicle}). 

In terms of network performance, the DSC of the segmentation path with a 0.72 mean on the SpineDepth dataset was comparable to \cite{felix2021towards} (DSC for tibia: 0.73), although segmentation of the spinal anatomy might be considered more challenging. The accuracy of the orientation prediction (16.33°) is comparable to the ones reported in the two publications inspiring our method (16.63° in \cite{mahendran20173d}, 13.59° in \cite{tulsiani2015viewpoints}).

Further analysis revealed a PCC of -0.78 between TRE and DSC, suggesting that the segmentation quality plays a key role in finding an accurate registration. The TRE also correlates (PCC of 0.74) with the visible bone surface error (VBSE) reported in the SpineDepth publication (\cite{liebmann2021spinedepth}), which essentially describes the reconstruction quality of the RGB-D sensor in use. While the dataset was recorded based on stereo calibrations created with a manufacturer-provided application, for our ex-vivo validation, standard OpenCV stereo calibration functionality (\cite{opencv_library}) was employed, leading to a much lower mean TRE (1.20 mm). This potential of accurate spine 3D reconstruction was confirmed in the study of \cite{manni2020towards}, where features in stereo grayscale images were matched with a 3D triangulation error below 0.5 mm. Stereo calibration and reconstruction quality may not be the only factors influencing the accuracy of the proposed registration approach, but they can be seen as a key factor. Other such factors could be the presence of soft tissue and the missing facet joints/mamillary processes in the dataset specimens, not only regarding accuracy, but also for convergence during general alignment, as more soft tissue flattens important bony surfaces, as well as the presence of previously inserted screws. The latter is suspected to be the reason for the increase in error from the first and second registration in the ex-vivo validation to the third, for which the specimen was rotated by 180° and all screws on the right side had already been inserted. This imbalance is not accounted for, which is a limitation of our method. More importantly, the full anatomical exposure in the cadaveric specimens is unrealistic within a clinical setting, and the high visibility facilitates the registration as well as the navigation. Furthermore, our method did not generalize to all specimens in the verification: specimen 10 had to be excluded due to its much smaller size compared to the other eight considered specimens.

For future work, the method needs to be evaluated on specimens with surgical approaches of varying sizes, i.e. less visibility of anatomical structures. Furthermore, transformer-based depth reconstruction, as proposed in \cite{gu2021calibration}, could be a promising way to increase registration accuracy, while feature-based tracking (\cite{manni2020towards}) should be investigated as a motion compensation strategy.

\section{Conclusions}
Our results suggest that fast, radiation-free, and fully automatic level-wise registration with real-time pose updates from RGB-D data for pedicle screw navigation under augmented reality guidance is feasible and meets clinical accuracy demands.

\section*{Declaration of Competing Interest}
The authors declare the following financial interests/personal relationships which may be considered as potential competing interests: Prof. Dr. med. Mazda Farshad, MPH is shareholder and member of the board of directors of Incremed AG, a company developing mixed-reality applications. All other authors declare that they have no conflict of interest.

\section*{Acknowledgments}
This project is part of SURGENT under the umbrella of University Medicine Zurich/Hochschulmedizin Zürich.

\bibliographystyle{unsrtnat}
\bibliography{references}

\end{document}